\documentclass{article}

\PassOptionsToPackage{sort,numbers, compress}{natbib}

 \usepackage[dblblindworkshop,final]{neurips_2025}
\workshoptitle{Machine Learning and the Physical Sciences (ML4PS)}

\usepackage[utf8]{inputenc} 
\usepackage[T1]{fontenc}    
\usepackage{hyperref}       
\usepackage{url}            
\usepackage{booktabs}       
\usepackage{amsfonts}       
\usepackage{nicefrac}       
\usepackage{microtype}      
\usepackage{xcolor}         
\usepackage{natbib}
\usepackage{algorithm}
\usepackage{algpseudocode}
\usepackage{multirow}
\usepackage[version=4]{mhchem}
\usepackage{graphicx}
\usepackage{amsmath}

\newcommand{\vecz}{\mathbf{z}}
\newcommand{\vecF}{\mathbf{F}}

\newcommand{\Samp}{\mathcal{S}}
\newcommand{\X}{\mathcal{X}}
\newcommand{\R}{\mathbf{R}}
\newcommand{\realspace}{\mathbb{R}}
\newcommand{\intspace}{\mathbb{Z}}

\title{P-DRUM: Post-hoc Descriptor-based Residual Uncertainty Modeling for Machine Learning Potentials}

\author{%
  Shih-Peng Huang\thanks{This work was conducted while the author was an intern at Preferred Networks. \\ Contact:  \url{nontawat@preferred.jp}, \url{tsuboi@preferred.jp}} \\
  Massachusetts Institute of Technology\\
  \texttt{sphuang@mit.edu} \\
  \And
  Nontawat Charoenphakdee \\
  Preferred Networks \\
  \texttt{nontawat@preferred.jp} \\
  \And
  Yuta Tsuboi \\
  Preferred Networks \\
  \texttt{tsuboi@preferred.jp} \\
  \And
  Yong-Bin Zhuang \\
  Preferred Networks \\
  \texttt{ybzhuang@preferred.jp} \\
  \And
  Wenwen Li \\
  Preferred Networks \\
  \texttt{wenwenli@preferred.jp} \\
}

\begin{document}

\maketitle

\begin{abstract}
Ensemble method is considered the gold standard for uncertainty quantification~(UQ) in machine learning interatomic potentials (MLIPs).
However, their high computational cost can limit its practicality. 
Alternative techniques, such as Monte Carlo dropout and deep kernel learning, have been proposed to improve computational efficiency; however, some of these methods cannot be applied to already trained models and may affect the prediction accuracy.
In this paper, we propose a simple and efficient post-hoc framework for UQ that leverages the descriptor of a trained graph neural network potential to estimate residual errors. 
We refer to this method as post-hoc descriptor-based residual uncertainty modeling (P-DRUM). 
P-DRUM models the discrepancy between MLIP predictions and ground truth values, allowing these residuals to act as proxies for prediction uncertainty. 
We explore multiple variants of P-DRUM and benchmark them against established UQ methods, evaluating both their effectiveness and limitations.
\end{abstract}

\section{Introduction}
Machine learning interatomic potentials (MLIPs) are transforming materials science by enabling atomic-scale simulations with the accuracy of quantum mechanical methods but at orders-of-magnitude higher computational efficiency~\citep{bpnet, gap, mtp, unke2021machine, behler2016perspective, deringer2019machine, oc20, batatia2022mace, takamoto2022towards, batatia2023foundation}. 
Despite their promise, the predictive reliability of MLIPs remains a critical concern, especially for atomic configurations outside the training data distribution~\citep{wollschlager2023uncertainty, tan2023single, kellner2024uncertainty, dai2025uncertainty}. 
Robust uncertainty quantification (UQ) is essential for assessing model reliability, guiding decision-making, and ensuring trustworthy simulation outcomes.

Several methods for UQ have been explored for MLIPs~\citep{janet2019quantitative,zhu2023fast,vita2025ltau}. 
Ensemble approaches, which aggregate predictions from multiple independently trained models, are widely regarded as the gold standard due to their effectiveness and their simplicity to implement~\citep{lakshminarayanan2017simple,tan2023single}.
However, ensembles are computationally expensive, particularly in settings with large-scale data or complex models like graph neural networks (GNN)~\citep{batatia2022mace, schutt2017schnet,dimenet, schutt2021equivariant}. 
Several alternative techniques have been proposed to address these limitations. 
Monte Carlo (MC) dropout~\citep{gal2016dropout} suggests to enable dropout during inference to introduce stochasticity into predictions at test time, and deep kernel learning combines neural networks with Gaussian processes~\citep{wollschlager2023uncertainty, rajput2023uncertainty, liu2020simple}. 
While these approaches improve efficiency, many of them require modification of the training pipeline or are incompatible with post-hoc applications, limiting their flexibility in scenarios where models are pre-trained and fixed.

Model descriptors have been shown to be effective in various downstream applications such as chemical property prediction~\citep{descriptor1, descriptor2}. 
Also, they have been combined with prediction errors for UQ, under the assumption that structures with higher prediction errors tend to exhibit greater uncertainty on average.
\citet{vita2025ltau} proposed loss trajectory analysis for uncertainty (LTAU), which leverages per-atom force error predictions to train an uncertainty model. 
While effective, this method requires logging the loss trajectory for every atom and is limited to capturing uncertainty in per-atom force predictions. 
In Orb-v3~\citep{rhodes2025orb}, prediction errors are discretized in a manner inspired by pLDDT in Alphafold~\citep{alphafold}, enabling joint optimization of the UQ objective alongside the prediction objective during model training.
In contrast to post-hoc methods, UQ objective of pLDDT is optimized jointly with the model training process.
For methods that are post-hoc that only utilizes model descriptor features,~\citet{janet2019quantitative} proposed to calculate the average feature distance between a test point and its k-nearest neighbors in the training data using revised autocorrelation descriptors. 
\citet{zhu2023fast} demonstrated the utility of fitting a Gaussian mixture model (GMM) for UQ in NequIP~\citep{nequip}. 
To the best of our knowledge, there has been limited study of post-hoc methods that explicitly estimate prediction errors using only the trained model, without requiring detailed training logs or specialized optimization procedures during training. 

In this paper, we explore post-hoc descriptor-based residual uncertainty modeling (P-DRUM) that utilize model descriptors of a trained graph neural network potential to estimate prediction error.
These residuals act as proxies for prediction error, enabling efficient and scalable UQ without modifying the original model architecture or training process. 
We investigate multiple variants of P-DRUM for energy and force errors: error-norm learning and deviation learning. 
Error-norm learning predicts a scalar representing the norm of the error, while deviation learning models the intrinsic error directly, maintaining the same format as the original prediction.

\section{Notation}
Let $X \in \X$ be a molecular or material structure in the input space $\X$.
To learn an MLIP, it is common that we use a training dataset with $N$ molecular or meterial structures: $\Samp =\{X_i, E_i, \vecF_i\}_{i=1}^{N}$.
Each structure $X_i$ consists of $n_i$ atoms and is represented as $X_i = \{\R_i, \vecz_i \}$. 
Here, $\R_i \in \realspace^{n_i \times 3}$ represents the atomic position information in three-dimensional space, 
and $\vecz_i \in \intspace^{n_i}$ encodes the atomic numbers corresponding to each atom within the structure.
For example, $z_{i1}= 1$ indicates that the first atom in structure $i$ is hydrogen.
$X_i$'s energy label $E_i \in \realspace$ and force label $\vecF_i \in \realspace^{n_i \times 3}$ are also provided for each structure.
With this training data, the goal of MLIP training is to accurately learn a real-valued function $f^{\mathrm{energy}}: \X \to \realspace$, which predicts energy $\widehat{E} \in \realspace$ given a structure $X=\{\R,\vecz\}$ of interest. 
Not only energy, we also expect that the force prediction is correctly predicted, where one can obtain the force information of atom $i$ given a predicted energy $E$ by calculating a negative gradient of $E$ (obtained via $f^{\mathrm{energy}}$) with respect to atomic position.
We note that some MLIPs directly predict forces (e.g., ForceNet~\citep{hu2021forcenet}), but these are beyond the scope of this paper.

\begin{figure*}
  \centering
  \includegraphics[width=1.0\linewidth]{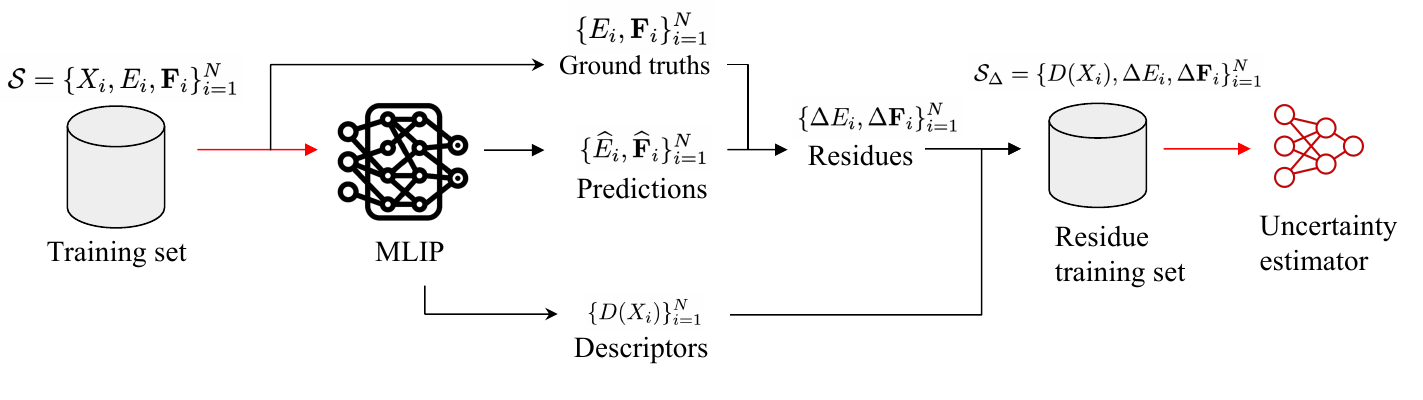} %
  \caption{Overview of the P-DRUM. Red arrow indicates ``model training using supervised learning''.}
  \label{fig:pdrum}
\end{figure*}

\section{Post-hoc descriptor-based residual uncertainty modeling (P-DRUM)}
Given structure $X = \{\R, \vecz\}$, in general, we can extract its descriptor in graph neural networks.
In this paper, we focus on a message-passing atomic cluster expansion (MACE)~\citep{batatia2022mace} model, where its descriptor (aka. features) can be extracted.
For $X$, we denote a function $D: \X \to \realspace^{d_{\mathrm{desc}} \times n }$, which maps a structure to a descriptor for each atom, where $d_{\mathrm{desc}}$ indicates the dimension of the descriptor.
we denote $D_{ij} \in \realspace^{d_{\mathrm{desc}}}$ a descriptor of atom $j$ in structure $i$.
Given a structure along with its energy and force $X$, $E$, $\vecF$, we can calculate energy residual $\Delta E=E-\widehat{E}$ and force residual $\Delta \vecF=\vecF-\widehat{\vecF}$, where $\widehat{E}$ and $\widehat{\vecF}$ indicate energy and force predictions of an MLIP. 
Given $N$ structures and a trained model, we can prepare the training data for P-DRUM: $\Samp_{\Delta} = \{D(X_i), \Delta E_i, \Delta \vecF_i \}_{i=1}^N$.
Using the residual training data $\Samp_{\Delta}$, we train a supervised model based on a multilayer perceptron (MLP) architecture, which offers significantly faster inference than MLIP models such as MACE.
Figure~\ref{fig:pdrum} shows the overview of P-DRUM.

\subsection{Energy residual learning} 
A naive approach for residual energy learning might involve designing a single model that processes an entire molecular structure with $n$ atoms and outputs the residual. 
However, this straightforward approach fails to naturally preserve permutational invariance and lacks the flexibility to handle systems of varying atomic sizes.
To address these challenges, we train a multilayered perceptron designed to take as input a descriptor for a single atom and output its corresponding scalar value $r^{\mathrm{s}}: \realspace^{d_{\mathrm{desc}}} \to \realspace$. 
Since the energy depends on the whole structure, but each structure can have different number of atoms, we model the energy residual as the sum of the atom-wise score function.
This approach allows the model to operate atom-wise, ensuring invariance properties and scalability to arbitrarily-sized structures.
In error norm learning, we can calculate a structure-wise squared loss by $\mathcal{L}_{\mathrm{E\text{-}norm}}(X_i) = ((\sum_j^{n_i} r_{\mathrm{E\text{-}norm}}^{\mathrm{s}}(D_{ij}) ) - |\Delta E_i|)^2$. 
For the energy deviation learning,
we learn $\Delta E$ directly, i.e., $\mathcal{L}_{\mathrm{E\text{-}diff}}(X_i) = ((\sum_j^{n_i} r_{\mathrm{E\text{-}diff}}^{\mathrm{s}}(D_{ij})) - \Delta E_i)^2$.



\subsection{Force residual learning}
Unlike energy, we designed a model that directly predicts force residuals using descriptors of individual atoms.
In error norm learning, we learn a function to estimate the Euclidean norm of the force error by minimizing the atomwise loss $\mathcal{L}_{\mathrm{F\text{-}norm}}(X_{ij}) = (r_{\mathrm{F\text{-}norm}}^{\mathrm{s}}(D_{ij})  - ||\Delta \vecF_{ij}||)^2$, where $r^{\mathrm{s}}_{\mathrm{F\text{-}norm}}$ is a real-valued function similarly to energy residual learning and $|| \cdot ||$ denotes the euclidean norm. 
In deviation learning, for each atom in a structure, we minimize the average of the coordinate-wise force-error loss: $\mathcal{L}_{\mathrm{F\text{-}diff}}(X_{ij}) = \frac{1}{3} \sum (r_{\mathrm{F\text{-}diff}}^{\mathrm{v}}(D_{ij})  - \Delta \vecF_{ij})^2$, where $r^{\mathrm{v}}_{\mathrm{F\text{-}diff}}: \realspace^{d_{\mathrm{desc}}} \to \realspace^3$ is a vector-valued function.
Uncertainty can be calculated by the Euclidean norm of the $3$-dimensional output of $r_{\mathrm{F\text{-}diff}}^{\mathrm{v}}$.
Similarly to energy residual learning, we use a multilayered perceptron to model the force residual function. 
\section{Experimental results}

\begin{table}
\centering
\caption{Dataset statistics used in this paper. 
HME21 consists of 37 different elements: H, Li, C, N, O, F, Na, Mg, Al, Si, P, S, Cl, K, Ca, Sc, Ti, V, Cr, Mn, Fe, Co, Ni, Cu, Zn, Mo, Ru, Rh, Pd, Ag, In, Sn, Ba, Ir, Pt, Au, and Pb.}
\begin{tabular}{lccccc}
\toprule
& \multicolumn{3}{c}{rMD17~\citep{rmd17}} & \ce{Ni3Al} & HME21~\citep{takamoto2022towards} \\
\cmidrule(lr){2-4}
& uracil & salicylic & malondialdehyde &  &  \\
\midrule
Elements & C,H,O,N & C,H,O & C,H,O & Ni, Al & 37 \\
Structure size (atom numbers) & 12 & 16 & 9 & 32 & 8--32 \\
Number of training data & 800 & 800 & 800 & 480 & 19956 \\
Number of validation data   & 200 & 200 & 200 & 120 & 2498 \\
Number of test data  & 1000 & 1000 & 1000 & 600 & 2495 \\
\bottomrule
\end{tabular}
\label{tab:dataset-stat}
\end{table}

\begin{table}
\centering
\caption{Five-trial average and standard deviation of Spearman correlation between prediction error and uncertainty of in-domain test data. 
The highest correlation values are highlighted in bold.}
\begin{tabular}{llcccccc} 
\toprule
\textbf{Error type} & \textbf{Method}   & \textbf{Uracil} & \textbf{Salicylic} & \textbf{Malondi-} & \textbf{\ce{Ni3Al}}  & \textbf{HME21} \\ 
&  & &  & \textbf{aldehyde} &  &  \\ 
\midrule
\multirow{6}{*}{\textbf{Energy}} 
                  & \textbf{Ensemble} & {0.04 (0.04)}            & {0.08 (0.04)}            & -0.01 (0.07)              & 0.39 (0.05)                     & {0.27 (0.02)}                     \\ 
                  & \textbf{MC-dropout}~\citep{gal2016dropout}  & -0.02 (0.02)            & -0.02 (0.02)            & -0.03 (0.02)              & -0.05 (0.05)                         & 0.20 (0.03)                     \\ 
                  & \textbf{GMM}~\citep{zhu2023fast}      & 0.07 (0.05)            & 0.07 (0.09)             & {0.13 (0.08)}              & {0.64 (0.05)}                        & 0.06 (0.03)                     \\ 
                  & \textbf{kNN}~\citep{janet2019quantitative}      & 0.06 (0.04)          & 0.06 (0.07)            & 0.09 (0.06)              & 0.64 (0.05)                     & -0.05 (0.03)                     \\                   
                  & \textbf{P-DRUM-norm}   & 0.12 (0.13)            & -0.01 (0.04)            & -0.09 (0.03)              & 0.62 (0.07)                    & \textbf{0.30 (0.03)}                     \\ 
                  & \textbf{P-DRUM-diff}    & \textbf{0.18 (0.14)}            & \textbf{0.16 (0.16)}            & \textbf{0.21 (0.14)}              & \textbf{0.87 (0.03)}                    & 0.26 (0.02)                     \\ 
\midrule
\midrule
\multirow{6}{*}{\textbf{Force}} 
                  & \textbf{Ensemble} & {\textbf{0.68 (0.01)}}            & 0.65 (0.01)            & 0.69 (0.01)              & {0.97 (0.00)}                         & 0.78 (0.00)                  \\
                  & \textbf{MC-dropout}~\citep{gal2016dropout}  & 0.24 (0.03)            & 0.27 (0.02)            & 0.27 (0.06)              & 0.87 (0.01)    & 0.68 (0.01)                     \\ 
                  & \textbf{GMM}~\citep{zhu2023fast}      & 0.58 (0.01)            & {0.67 (0.03)}            & {0.68 (0.02)}              & 0.96 (0.01)                & 0.64 (0.04)                     \\ 
                  & \textbf{kNN}~\citep{janet2019quantitative}      & 0.52 (0.02)            & 0.61 (0.04)            & 0.65 (0.02)              & 0.96 (0.01)               & 0.54 (0.01)                     \\                   
                  & \textbf{P-DRUM-norm}   & 0.67 (0.02)            & \textbf{0.71 (0.04)}            & \textbf{0.69 (0.02)}              & \textbf{0.98 (0.00)}                     & \textbf{0.92 (0.00)}                     \\ 
                  & \textbf{P-DRUM-diff}    & 0.53 (0.02)            & 0.58 (0.04)            & 0.57 (0.01)              & 0.96 (0.01)                    & {0.85 (0.01)}                     \\ 
\bottomrule
\end{tabular}
\label{tab:id-spearman}
\end{table}


We compare P-DRUM against ensemble, MC-dropout~\citep{gal2016dropout}, k-nearest neighbors of descriptors proposed by~\citet{janet2019quantitative}, and GMM of descriptors proposed by~\citet{zhu2023fast}. 
We used HME21 dataset which contains 37 elements~\citep{takamoto2022towards}, and three datasets from rMD17 datasets~\citep{rmd17}: malondialdehyde, salicylic acid, uracil.
Table~\ref{tab:dataset-stat} shows statistics of all datasets used in this paper. 
Additionally, we assess out-of-distribution (OOD) robustness with a benchmark on the nickel aluminide (\ce{Ni3Al}) dataset, generated using Matlantis~\citep{matlantis}.
We obtained the MACE model from the official MACE repository~\footnote{\url{https://github.com/ACEsuit/mace}} and used its implementation of the descriptor.
Since MACE does not support MC-dropout by default, we implemented dropout in the fully connected layers after the activation function of both the interaction block and the readout block.
Additional training details for our experiments are provided in Appendix~\ref{app:hyperparameters}.
In terms of computational efficiency, ensemble and MC-dropout require five forward passes of MACE, whereas kNN, GMM, and P-DRUM need only a single pass, with only negligible additional overhead compared to MACE forward pass cost.

\subsection{Uncertainty-error correlation evaluation}
In this section, we evaluate how uncertainty scores relate to prediction errors using Spearman correlation. 
This evaluation is based on the idea that structures with higher uncertainty should tend to exhibit larger errors.
We analyze both energy and force errors in our experiments. 
Note that only the kNN and GMM uncertainties do not differentiate between energy and force uncertainties. 
As a result, we used the same uncertainty estimates for evaluating both energy and force performance for them.

Table~\ref{tab:id-spearman} reports the 5-trial average Spearman correlation for each method across datasets.
The results indicate that P-DRUM-diff performs relatively well in predicting energy uncertainties, whereas P-DRUM-norm is more effective for force uncertainties. 
Notably, the performance gap between GMM and kNN is larger in HME21. 
For readers who are interested, we provide an additional analysis using principal component analysis of HME21 in Appendix~\ref{app:P-DRUM-vs-gmm}.  

\subsection{OOD detection of \ce{Ni3Al}}
\label{subsec:ood-ni3al}
In this section, we evaluate the performance in the task of out-of-distribution (OOD) detection.
We used the \ce{Ni3Al} dataset, which was generated using Preferred Potential (PFP)~\citep{takamoto2022towards} within the Matlantis platform~\citep{matlantis}, where the initial structure is collected from Materials Project (mp-2593)~\footnote{https://next-gen.materialsproject.org/materials/mp-2593}~\citep{jain10materials}. 
To assess OOD detection, we prepared several distinct OOD datasets for \ce{Ni3Al}: 
(1) High-temperature OOD: structures generated via molecular dynamics simulations at temperatures higher than those used in the training dataset. 
The training data included temperatures of 500K, 1000K, and 1500K, while the OOD data was derived from simulations at 2000K and 3000K. 
(2) Hexagonal: \ce{Ni3Al} with different phase from original dataset (mp-1183232).
(3) Cubic: \ce{Ni3Al} with different phase from the original dataset (mp-672232).
(4) Swap: randomly swap positions of Ni and Al in the structures for 2, 4, and 8 pairs. 
We used PFP predictions as ground truths and force uncertainty for evaluation.

Table~\ref{tab:ood-combined-force} summarizes the performance of various OOD detection methods. 
Ensemble, kNN, GMM, and P-DRUM-diff consistently achieve strong OOD detection performance across the benchmark.
In comparison, P-DRUM-norm performs less effectively, while MC-dropout yields the lowest performance in terms of both Spearman correlation and AUC.
Although P-DRUM-norm achieves the best performance in Table~\ref{tab:id-spearman}, it proves less effective in the OOD setting in our experiments.

\begin{table}
\centering
\caption{Five-trial average of Spearman correlation and AUC performance in \ce{Ni3Al} OOD detection suite for each method. Standard deviation is omitted due to space constraint. \textbf{Force uncertainty} is used for ensemble, dropout, and P-DRUM.}
\label{tab:ood-combined-force}
\begin{tabular}{lcc|cc|cc|cc|cc}
\toprule
\textbf{Method} & \multicolumn{2}{c}{\textbf{High temp.}} & \multicolumn{2}{c}{\textbf{Hexagonal}} & \multicolumn{2}{c}{\textbf{Cubic}} & \multicolumn{2}{c}{\textbf{Swap}} & \multicolumn{2}{c}{\textbf{All}} \\ 
\cmidrule(lr){2-3} \cmidrule(lr){4-5} \cmidrule(lr){6-7} \cmidrule(lr){8-9} \cmidrule(lr){10-11}
        & Corr. & AUC & Corr. & AUC & Corr. & AUC & Corr. & AUC & Corr. & AUC \\
\midrule
\textbf{Ensemble}   & 0.98 & \textbf{1.00} & 0.88 & 0.94 & \textbf{0.95} & \textbf{1.00} & 0.76 & \textbf{1.00} & \textbf{0.90} & 0.99 \\
\textbf{MC-dropout}~\citep{gal2016dropout}    & 0.92 & \textbf{1.00} & 0.54 & 0.63 & 0.81 & 0.84 & 0.62 & 0.82 & 0.72 & 0.82 \\
\textbf{GMM}~\citep{zhu2023fast}        & 0.98 & \textbf{1.00} & 0.74 &\textbf{1.00} & 0.73 & \textbf{1.00} & 0.77 & \textbf{1.00} & 0.81 & \textbf{1.00} \\
\textbf{kNN}~\citep{janet2019quantitative}        & 0.98 & \textbf{1.00} & 0.80 & 0.99 & 0.75 & \textbf{1.00} & 0.76 & \textbf{1.00} & 0.82 & \textbf{1.00}\\
\textbf{P-DRUM-norm}  & \textbf{0.99} & \textbf{1.00} & 0.82 & 0.82 & 0.78 & 0.82 & 0.70 & 0.99 & 0.82 & 0.91 \\
\textbf{P-DRUM-diff}   & 0.97 & \textbf{1.00} & \textbf{0.89} & 0.97 & 0.80 & \textbf{1.00} & \textbf{0.81} & \textbf{1.00} & 0.87 & 0.99 \\
\bottomrule
\end{tabular}
\end{table}

\section{Discussions}
\label{sec:discussion}

\textbf{When P-DRUM outperform kNN, GMM?}
While descriptor-based methods perform competitively, P-DRUM clearly outperforms kNN and GMM in error–uncertainty correlation on the HME21 dataset. 
We hypothesize that when a dataset contains many elements, descriptor information alone may be difficult to capture the error correlation, and incorporating the prediction error signal can enhance the uncertainty estimation by aligning it more closely with the true prediction error.

\textbf{Which approach is better: error norm learning or deviation learning?}
In Table~\ref{tab:id-spearman}, deviation learning outperforms error-norm learning for energy Spearman correlation, whereas the opposite trend is observed for force correlation. 
This difference likely arises from the nature of the targets: energy deviation is a scalar, where retaining the error sign aids learning, while force deviation is a three-dimensional vector, making direct estimation more challenging. 
Using the force error norm reduces this complexity, resulting in an improvement in the Spearman correlation.
However, as shown in Table~\ref{tab:ood-combined-force}, using force uncertainty, error-norm learning underperforms deviation learning in OOD detection, indicating the need for further investigation into the advantages and limitations of these approaches. 
We hypothesize that compressing errors ($3$-dimensional for forces) into a norm ($1$-dimensional) might be detrimental for OOD detection.

\section{Conclusion and future work}
We investigated the effectiveness of post-hoc descriptor-based residual uncertainty modeling (P-DRUM) for machine learning interatomic potentials (MLIPs).
P-DRUM achieved a higher Spearman correlation with prediction errors compared to other methods.
However, P-DRUM-norm showed inferior out-of-distribution (OOD) detection performance compared to Gaussian mixture model (GMM) and k-nearest neighbor (kNN) descriptor-based approaches, revealing its potential limitations.
Future work will explore extending P-DRUM to active learning pipelines and broader MLIP applications to further evaluate its versatility, as well as developing improved training strategies to enhance its performance.
In addition, assessing the reliability of reusing the training set to construct the residual training set is an important direction.
Using separate dataset splits for training the MLIP and the uncertainty estimation model can reduce bias from the original data but also limits the available training samples. 
Exploring this trade-off is useful to developing a data-efficient P-DRUM strategy.



\begin{ack}
    We would like to thank Professor Ju Li for helpful comments and suggestions.
\end{ack}

\bibliography{references}






\appendix

\begin{table}
\centering
\caption{Five-trial average and standard deviation of AUC using uncertainty to classify low error and high error class on in-domain test data. 
The highest correlation values are highlighted in bold.}
\begin{tabular}{llcccccc} 
\toprule
\textbf{Error type} & \textbf{Method}   & \textbf{Uracil} & \textbf{Salicylic} & \textbf{Malondialdehyde} & \textbf{\ce{Ni3Al}}  & \textbf{HME21} \\ 
\midrule
\multirow{6}{*}{\textbf{Energy}} 
                  & \textbf{Ensemble}                           & {0.48 (0.02)}          & {0.53 (0.02)}          & 0.50 (0.03)           & 0.58 (0.03)                 & {0.58 (0.01)}                     \\ 
                  & \textbf{MC-dropout}~\citep{gal2016dropout}     & -0.48 (0.01)           & 0.49 (0.02)           & 0.50 (0.01)           & 0.49 (0.04)                & 0.55 (0.02)                     \\ 
                  & \textbf{GMM}~\citep{zhu2023fast}            & 0.47 (0.06)            & 0.46 (0.07)            & {0.55 (0.04)}          & {0.71 (0.04)}               & 0.54 (0.01)                     \\ 
                  & \textbf{kNN}~\citep{janet2019quantitative}  & 0.48 (0.05)            & 0.47 (0.05)            & 0.54 (0.03)            & 0.70 (0.04)                 & 0.51 (0.01)                     \\ 
                  & \textbf{P-DRUM-norm}                          & 0.51 (0.03)            & 0.48 (0.01)           & 0.46 (0.02)           & 0.72 (0.03)                 & \textbf{0.63 (0.02)}                     \\ 
                  & \textbf{P-DRUM-diff}                          & \textbf{0.56 (0.07)}   & \textbf{0.57 (0.06)}   & \textbf{0.59 (0.06)}   & \textbf{0.90 (0.04)}        & 0.61 (0.01)                     \\ 
\midrule
\midrule
\multirow{6}{*}{\textbf{Force}} 
                  & \textbf{Ensemble}                           & {0.83 (0.01)}          & {0.83 (0.01)}         & 0.83 (0.01)           & \textbf{0.98 (0.00)}         & {0.93 (0.00)}                     \\ 
                  & \textbf{MC-dropout}~\citep{gal2016dropout}     & 0.62 (0.03)            & 0.63 (0.02)           & 0.63 (0.04)           & 0.95 (0.01)                 & 0.83 (0.01)                     \\ 
                  & \textbf{GMM}~\citep{zhu2023fast}            & 0.79 (0.00)   & 0.84 (0.02)           & {0.82 (0.02)}         & 0.98 (0.01)                 & 0.85 (0.03)                     \\ 
                  & \textbf{kNN}~\citep{janet2019quantitative}  & 0.77 (0.01)            & 0.81 (0.02)           & 0.81 (0.02)           & 0.98 (0.01)                 & 0.84 (0.00)                     \\ 
                  & \textbf{P-DRUM-norm}                          & \textbf{0.85 (0.01)}            & \textbf{0.86 (0.03)}  & \textbf{0.83 (0.01)}  & \textbf{0.98 (0.00)}        & \textbf{0.98 (0.00)}                     \\ 
                  & \textbf{P-DRUM-diff}                          & 0.76 (0.01)            & 0.78 (0.02)           & 0.77 (0.02)           & 0.96 (0.06)                 & 0.92 (0.01)                     \\ 
\bottomrule
\end{tabular}
\label{tab:id-AUC}
\end{table}

\section{Broader Impact}
\label{app:broader-impact}
This work focuses on developing uncertainty estimation techniques called post-hoc descriptor-based residual uncertainty modeling (P-DRUM) for machine learning interatomic potentials. 
P-DRUM is designed to improve the reliability and robustness of simulations in chemistry, materials science, and related fields. 
Importantly, our research does not involve human subjects, personal data, or any form of unethical experimentation. 
The techniques we propose are purely computational and are evaluated on standard benchmark datasets and simulated molecular systems.

That said, as with many advances in machine learning and computational modeling, there is the possibility that the methods we develop could be misused. 
More accurate and reliable atomistic simulations may be applied in contexts that could lead to harmful outcomes, for instance in the design of materials for military applications or environmentally damaging technologies. 
We strongly discourage the use of our methods in ways that could contribute to unethical purposes.


\section{Training details and hyperparameters}
\label{app:hyperparameters}
All models were trained using the default MACE architecture with 32 channels, a radial cutoff of 5 \AA, a batch size of 50, and two interaction layers (RealAgnosticInteractionBlock and RealAgnosticInteractionResidualBlock), yielding 64 descriptor dimensions. Training ran for 100 epochs: for the first 75 epochs, energy and force loss weights were 1 and 100, respectively; for the final 25 epochs, the energy weight was increased to 1000, with the force weight unchanged.

During the P-DRUM experiments, validation set was used for learning rate scheduling and early stopping. Training began with a learning rate of $10^{-3}$ and patience of 10 epochs, halving the learning rate whenever the validation error failed to improve for 10 consecutive epochs. Training was terminated after a maximum of 1000 epochs or once the learning rate decreased to $10^{-7}$. 

Except for \ce{Ni3Al} and HME21 dataset, in which we used a batch size of $2048$ for P-DRUM-diff forces prediction, the batch size was set to $64$ atoms in all the other P-DRUM forces training. For P-DRUM energy training, the batch size was set to $64$ chemical structures. 
The larger dataset sizes in these cases lead to a greater number of total atoms per batch, and increasing the batch size helped stabilize training. 
For P-DRUM-norm energy and forces prediction, we employed the ReLU activation function with a single hidden layer, along with a softplus activation right before output. P-DRUM-diff uses similar but softplus-removed MLP architecture, where one hidden layer was used for energy and two hidden layers were used for forces.
For MC-dropout, we set the dropout ratio to 10\%.

For computing resources, we used an NVIDIA V100 GPU (32 GB) of memory for training different trials. The execution time depends on the dataset. 
Although we did not precisely measure the training time, each trial of each method can be completed within 7 hours on a single GPU, including training a MACE model and uncertainty estimation method.

\section{Additional experimental results}
\subsection{AUC evaluation of in-domain dataset}
Here, we show the results of AUC where we split test data into two classes: low error and high error classes.
We put lowest 20\% error as low error class and high error otherwise.
Table~\ref{tab:id-AUC} shows the comparison of the AUC performance across all different methods.

\subsection{OOD detection results with standard deviation}
\label{app:add-exp}
Table~\ref{tab:ood-combined-force-spearman} shows the spearman correlation comparisons and Table~\ref{tab:ood-combined-force-AUC} shows the AUC comparisons.

\begin{table}[t]
\centering
\caption{Five-trial average and standard error of Spearman correlation performance in \ce{Ni3Al} OOD detection suite for each method. \textbf{Force uncertainty} is used for ensemble, dropout, and P-DRUM.}
\begin{tabular}{lccccc}
\toprule
\textbf{Method}   & \textbf{High temp.} & \textbf{Hexagonal} & \textbf{Cubic} & \textbf{Swap}  & \textbf{Average} \\ 
\midrule
\textbf{Ensemble}                           & 0.98 (0.00)  & 0.88 (0.01)  & \textbf{0.95 (0.01)}  & 0.76 (0.01)  & \textbf{0.90} \\
\textbf{MC-dropout}~\citep{gal2016dropout}    & 0.92 (0.01)  & 0.54 (0.10) & 0.81 (0.03)  & 0.62 (0.06) & 0.72  \\
\textbf{GMM}~\citep{zhu2023fast}        & 0.98 (0.00)  & 0.74 (0.05) & 0.73(0.02)  & 0.77 (0.04)  & 0.81\\
\textbf{kNN}~\citep{janet2019quantitative}   & 0.98 (0.01) & 0.80 (0.03)  & 0.75 (0.03) &  0.76 (0.07) & 0.82  \\
\textbf{P-DRUM-norm}  & \textbf{0.99 (0.00)} & 0.82 (0.10)  & 0.78 (0.26)  & 0.70 (0.06)  & 0.82 \\
\textbf{P-DRUM-diff}   & 0.97 (0.00) & \textbf{0.89 (0.04)} & 0.80 (0.08) & \textbf{0.81 (0.03)} & 0.87 \\
\bottomrule
\end{tabular}
\label{tab:ood-combined-force-spearman}
\end{table}

\begin{table}[t]
\centering
\caption{Five-trial average and standard error of AUC performance in \ce{Ni3Al} OOD detection suite for each method. \textbf{Force uncertainty} is used for ensemble, dropout, and P-DRUM.}
\begin{tabular}{lccccc}
\toprule
\textbf{Method}   & \textbf{High temp.} & \textbf{Hexagonal} & \textbf{Cubic} & \textbf{Swap}  & \textbf{Average} \\ 
\midrule
\textbf{Ensemble}                            & \textbf{1.00 (0.00)}  & 0.94 (0.02) & \textbf{1.00 (0.00)} & \textbf{1.00 (0.00)}  & 0.99 \\
\textbf{MC-dropout}~\citep{gal2016dropout}    & \textbf{1.00 (0.00)} & 0.63 (0.04)  & 0.84 (0.04)  & 0.82 (0.08)  & 0.82 \\
\textbf{GMM}~\citep{zhu2023fast}        & \textbf{1.00 (0.00)}  &\textbf{1.00 (0.00)} & \textbf{1.00 (0.00)} & \textbf{1.00 (0.00)}& \textbf{1.00} \\
\textbf{kNN}~\citep{janet2019quantitative}  & \textbf{1.00 (0.00)}  & 0.99 (0.01)  & \textbf{1.00 (0.00)}  & \textbf{1.00 (0.00)}  & \textbf{1.00}\\
\textbf{P-DRUM-norm}   & \textbf{1.00 (0.00)}  & 0.82 (0.09)  & 0.82 (0.25) & 0.99 (0.02)  & 0.91 \\
\textbf{P-DRUM-diff}   & \textbf{1.00 (0.00)}  & 0.97 (0.04) & \textbf{1.00 (0.00)} & \textbf{1.00 (0.00)} & 0.99 \\
\bottomrule
\end{tabular}
\label{tab:ood-combined-force-AUC}
\end{table}

\section{When does P-DRUM outperform kNN and GMM?: an analysis based on principle component analysis (PCA)}
\label{app:P-DRUM-vs-gmm}
\begin{figure}[t]
    \centering
    \includegraphics[width=0.9\textwidth]{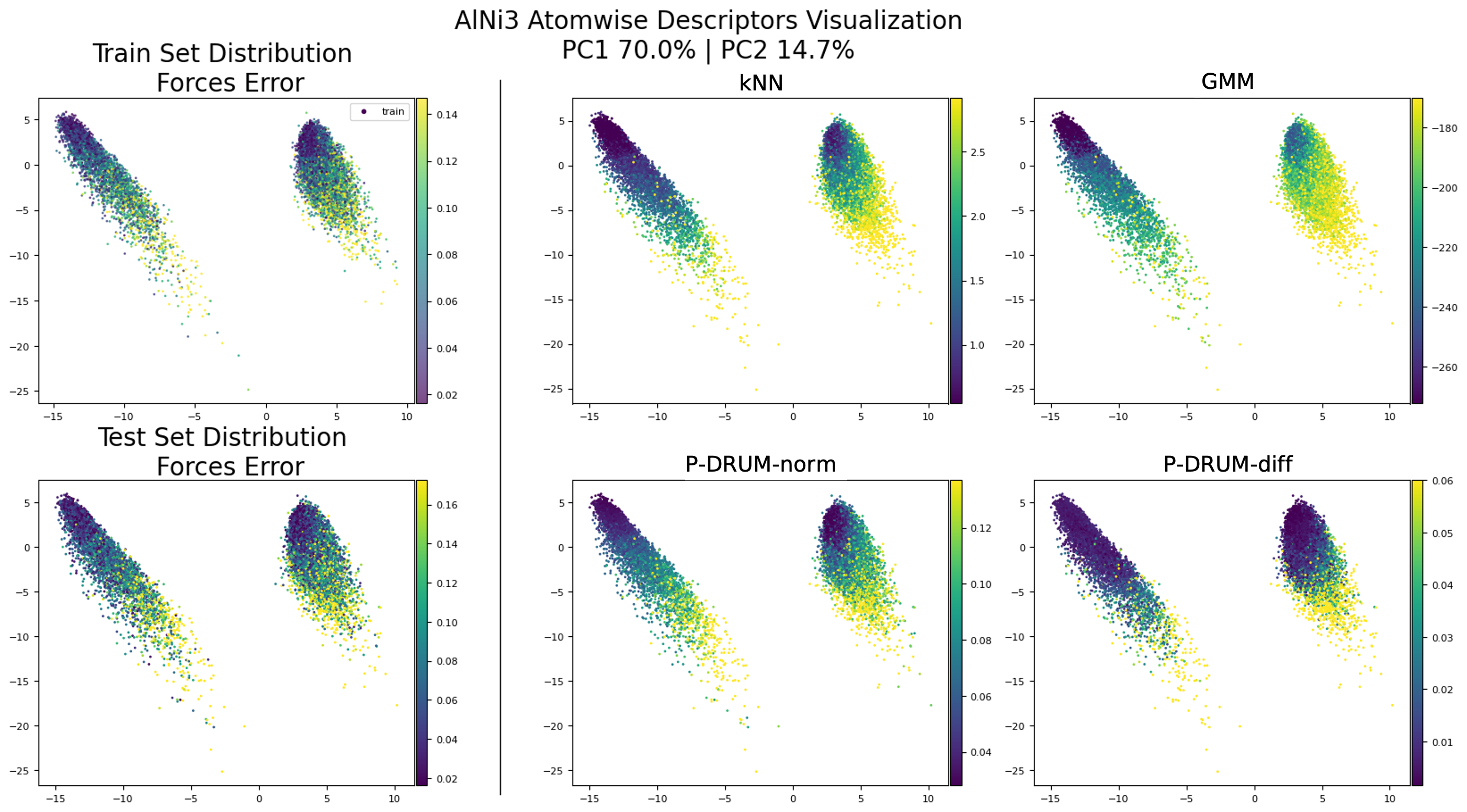}
    \caption{
        PCA visualization of the \ce{Ni3Al} dataset.
        The left subplots show the prediction error of train and test set in PC space, while the uncertainty metrics of the test set on the right subplots.
    }
    \label{fig:AlNi3_PCA_visualization}
\end{figure}

In the in-domain setup, we observed that the P-DRUM-norm method performed the best across all methods; however, the baseline descriptor methods (kNN and GMM) also performed comparably on almost all datasets except HME21. Taking the \ce{Ni3Al} dataset as an example (Figure~\ref{fig:AlNi3_PCA_visualization}), the descriptors in PC space for each atom in the train and test set were plotted, and each cluster in the plot represents the atoms of Ni or Al. The lower force error atoms in the figure represent atoms with lower force prediction error or low uncertainty. In the training set distribution, denser regions on the PCA plot correspond to lower force errors, i.e., smaller error magnitudes. This indicates that descriptor-based baseline methods such as GMM and kNN can achieve good performance without requiring prior knowledge of the force errors.

\begin{figure}[t]
    \centering
    \includegraphics[width=0.9\textwidth]{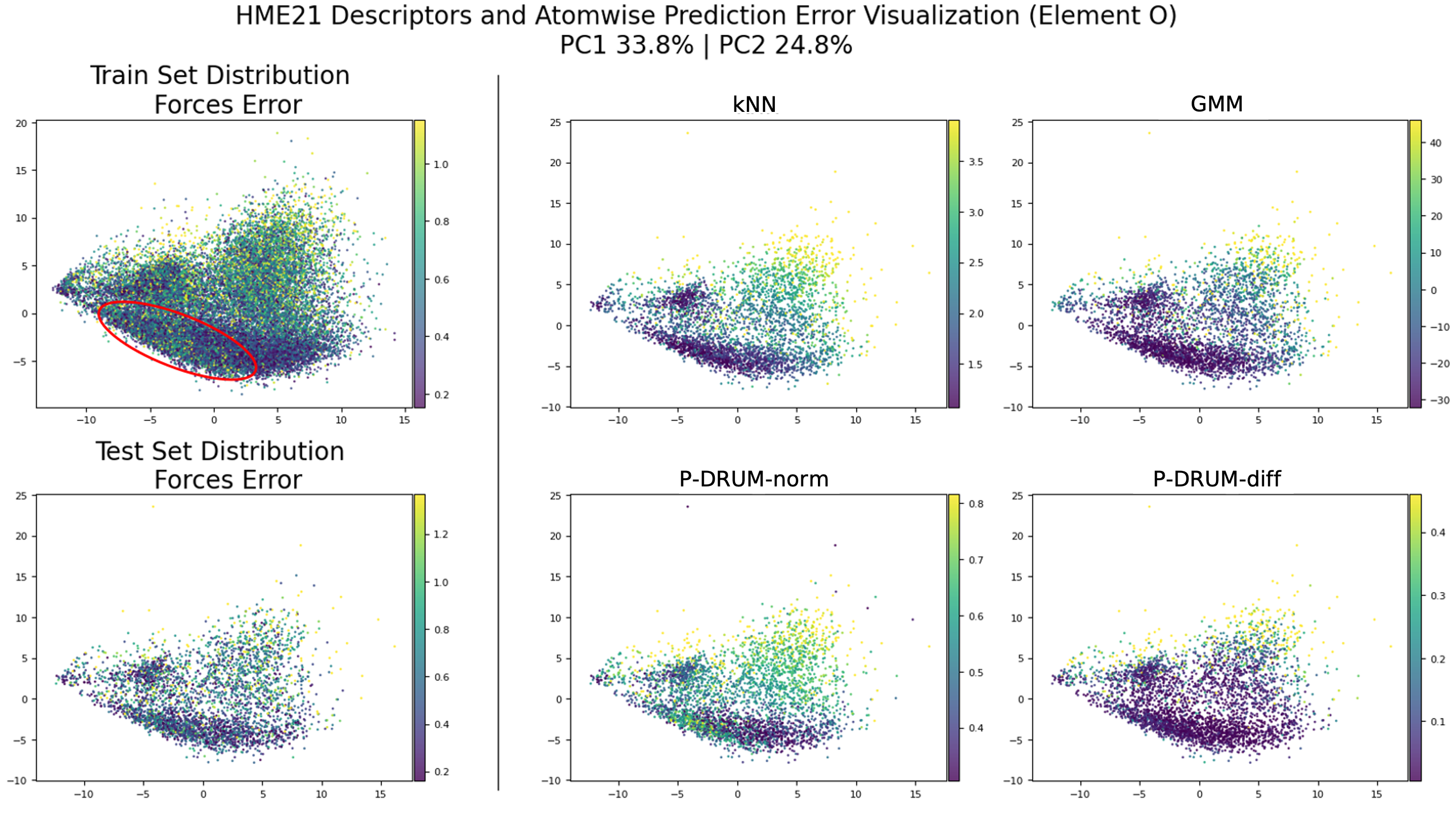}
    \caption{
        PCA visualization of the oxygen atoms in HME21 dataset.
        The left subplots show the prediction error of train and test set in PC space, while the uncertainty metrics of the test set on the right subplots.
    }
    \label{fig:HME21_PCA_visualization_O_only}
\end{figure}

\begin{figure}[t]
    \centering
    \includegraphics[width=0.9\textwidth]{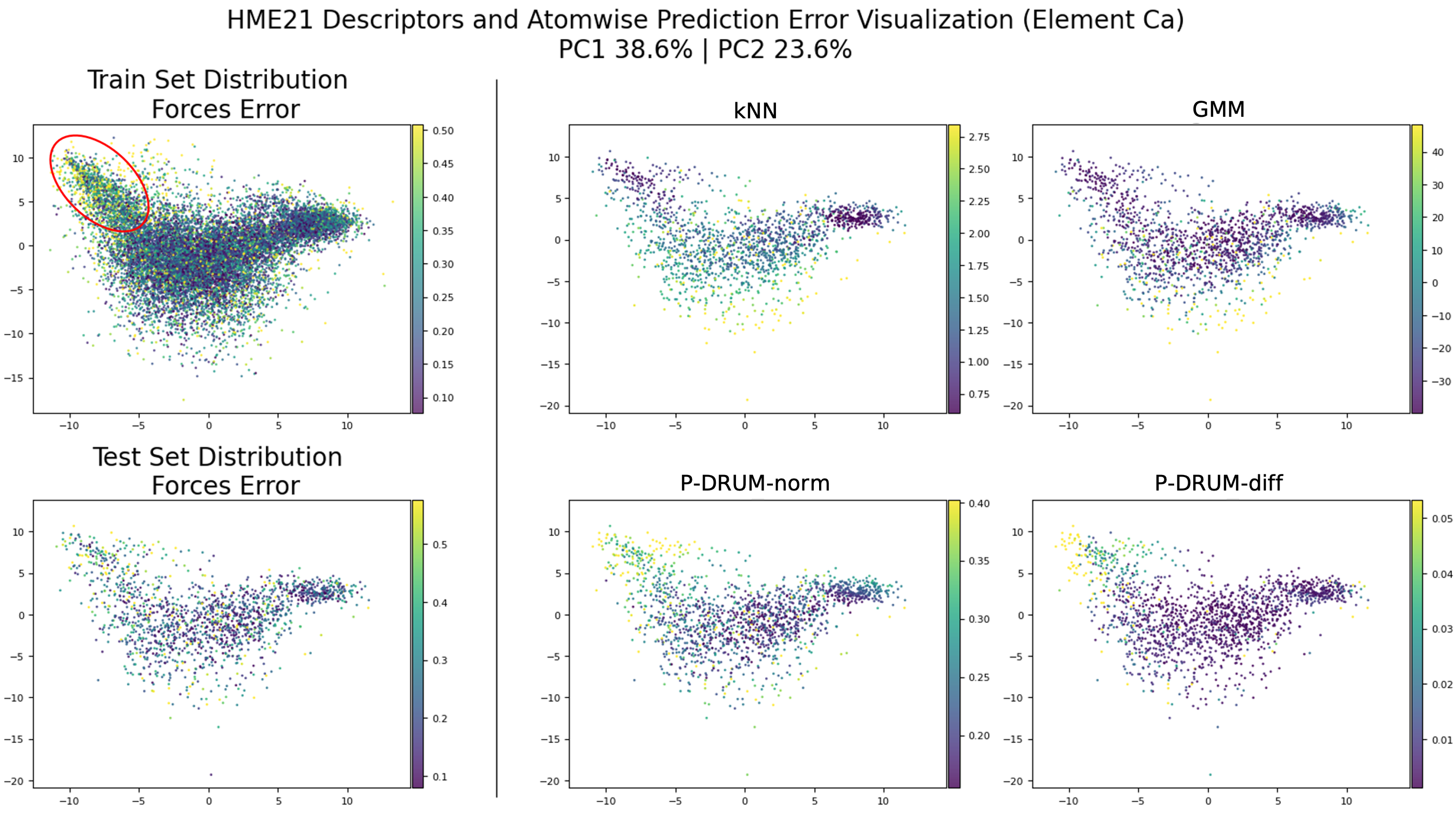}
    \caption{
        PCA visualization of the calcium atoms in HME21 dataset.
        The left subplots show the prediction error of train and test set in PC space, while the uncertainty metrics of the test set on the right subplots.
    }
    \label{fig:HME21_PCA_visualization_Ca_only}
\end{figure}

However, comparing to rMD17 elements and \ce{Ni3Al} that contain at most 4 types of elements in each dataset, the HME21 dataset contains 37 different elements and a more diverse interaction between different elements. The landscape of prediction error is more complicated and difficult to learn without having the error information of the training set. Figure~\ref{fig:HME21_PCA_visualization_O_only} shows the oxygen atoms, which is the most common atom in HME21 descriptors in PC space. We observed that in the train set, the circled area is the densest while having slightly higher force prediction error than the area on the right of the circle. kNN and GMM was unable to capture this and predicted the circled area as lowest uncertainty, while P-DRUM-norm method learned this from the forces prediction error during the training step. 

Similar trends are observed in Figure~\ref{fig:HME21_PCA_visualization_Ca_only}, which only shows the calcium atoms in HME21. The top‑left region of the PCA plot is densely populated, yet exhibits relatively high force‑prediction errors. Both of our P-DRUM methods successfully capture this behavior, as the prediction error was explicitly incorporated into the training process. In contrast, the kNN and GMM approaches rely solely on descriptor information and therefore incorrectly assign low uncertainty to the same tail region in the top left. These results suggest that P-DRUM methods have strong potential for uncertainty prediction in more diverse and complex datasets, paving the way toward the development of universal potentials uncertainty estimation.

\newpage
\section*{NeurIPS Paper Checklist}

\begin{enumerate}

\item {\bf Claims}
    \item[] Question: Do the main claims made in the abstract and introduction accurately reflect the paper's contributions and scope?
    \item[] Answer: \answerYes{} 
    \item[] Justification: We propose P-DRUM and evaluate its performance to show the capability of our method.

\item {\bf Limitations}
    \item[] Question: Does the paper discuss the limitations of the work performed by the authors?
    \item[] Answer: \answerYes{} 
    \item[] Justification: We discussed limitations of our method in the conclusion and future work in the Section~\ref{sec:discussion}.

\item {\bf Theory assumptions and proofs}
    \item[] Question: For each theoretical result, does the paper provide the full set of assumptions and a complete (and correct) proof?
    \item[] Answer: \answerNA{} 
    \item[] Justification: \answerNA{}

    \item {\bf Experimental result reproducibility}
    \item[] Question: Does the paper fully disclose all the information needed to reproduce the main experimental results of the paper to the extent that it affects the main claims and/or conclusions of the paper (regardless of whether the code and data are provided or not)?
    \item[] Answer: \answerYes{} 
    \item[] Justification: We share the training information and disclose training hyperparameters in Appendix~\ref{app:hyperparameters}.

\item {\bf Open access to data and code}
    \item[] Question: Does the paper provide open access to the data and code, with sufficient instructions to faithfully reproduce the main experimental results, as described in supplemental material?
    \item[] Answer: \answerNo{} 
    \item[] Justification: We do not plan to release the code for the workshop submission version.

\item {\bf Experimental setting/details}
    \item[] Question: Does the paper specify all the training and test details (e.g., data splits, hyperparameters, how they were chosen, type of optimizer, etc.) necessary to understand the results?
    \item[] Answer: \answerYes{} 
    \item[] Justification: We listed hyperparameters in Appendix~\ref{app:hyperparameters} and describe training dataset in the main body. However, we also used \ce{Ni3Al} dataset which is not open source but we believe we provided enough information for readers to understand the results.

\item {\bf Experiment statistical significance}
    \item[] Question: Does the paper report error bars suitably and correctly defined or other appropriate information about the statistical significance of the experiments?
    \item[] Answer: \answerYes{} 
    \item[] Justification: We conducted experiments for 5 trials to produce mean and standard deviation of the results.

\item {\bf Experiments compute resources}
    \item[] Question: For each experiment, does the paper provide sufficient information on the computer resources (type of compute workers, memory, time of execution) needed to reproduce the experiments?
    \item[] Answer: \answerYes{} 
    \item[] Justification: We described this information in Appendix~\ref{app:hyperparameters}.
    
\item {\bf Code of ethics}
    \item[] Question: Does the research conducted in the paper conform, in every respect, with the NeurIPS Code of Ethics \url{https://neurips.cc/public/EthicsGuidelines}?
    \item[] Answer: \answerYes{} 
    \item[] Justification: We have read the code of ethics and believe our research conforms to this NeurIPS Code of Ethics.

\item {\bf Broader impacts}
    \item[] Question: Does the paper discuss both potential positive societal impacts and negative societal impacts of the work performed?
    \item[] Answer: \answerYes{} 
    \item[] Justification: We wrote a broader impact section in Appendix~\ref{app:broader-impact}.
    
\item {\bf Safeguards}
    \item[] Question: Does the paper describe safeguards that have been put in place for responsible release of data or models that have a high risk for misuse (e.g., pretrained language models, image generators, or scraped datasets)?
    \item[] Answer: \answerNA{} 
    \item[] Justification: \answerNA{}

\item {\bf Licenses for existing assets}
    \item[] Question: Are the creators or original owners of assets (e.g., code, data, models), used in the paper, properly credited and are the license and terms of use explicitly mentioned and properly respected?
    \item[] Answer: \answerYes{} 
    \item[] Justification: We cite the source of datasets and code of the MACE repository in the paper.

\item {\bf New assets}
    \item[] Question: Are new assets introduced in the paper well documented and is the documentation provided alongside the assets?
    \item[] Answer: \answerYes{} 
    \item[] Justification: We provided information of how \ce{Ni3Al} dataset was generated in the experiment section.

\item {\bf Crowdsourcing and research with human subjects}
    \item[] Question: For crowdsourcing experiments and research with human subjects, does the paper include the full text of instructions given to participants and screenshots, if applicable, as well as details about compensation (if any)? 
    \item[] Answer: \answerNA{} 
    \item[] Justification: \answerNA{}

\item {\bf Institutional review board (IRB) approvals or equivalent for research with human subjects}
    \item[] Question: Does the paper describe potential risks incurred by study participants, whether such risks were disclosed to the subjects, and whether Institutional Review Board (IRB) approvals (or an equivalent approval/review based on the requirements of your country or institution) were obtained?
    \item[] Answer: \answerNA{} 
    \item[] Justification: The paper does not involve crowdsourcing nor research with human subjects

\item {\bf Declaration of LLM usage}
    \item[] Question: Does the paper describe the usage of LLMs if it is an important, original, or non-standard component of the core methods in this research? Note that if the LLM is used only for writing, editing, or formatting purposes and does not impact the core methodology, scientific rigorousness, or originality of the research, declaration is not required.
    \item[] Answer: \answerNA{} 
    \item[] Justification: The core method development in this research does not involve LLMs as any important, original, or non-standard components.

\end{enumerate}

\end{document}